\documentclass{article}

\usepackage{microtype}
\usepackage{graphicx}
\usepackage{booktabs} 

\PassOptionsToPackage{numbers}{natbib}

\usepackage{microtype}

\usepackage{xurl, hyperref}
\usepackage[super]{nth}
\usepackage[ruled]{algorithm2e}
\usepackage{float, graphicx, wrapfig, subcaption}
\usepackage{booktabs, tabularx, multirow, tablefootnote}
\usepackage{listings}
\usepackage{nicefrac}
\usepackage{amsmath, amsthm}

\usepackage{microtype}
\usepackage{graphicx}
\usepackage{comment}
\usepackage{color}
\usepackage{colortbl}
\usepackage{xcolor}
\usepackage{colortbl}
\usepackage{hyphenat}
\usepackage{amsmath}
\usepackage{tabularx}
\usepackage{algorithmic}
\usepackage{enumitem}
\usepackage{mathrsfs}
\usepackage[justification=centering]{caption}

\usepackage{needspace}

\usepackage{svg}
\usepackage{hyperref}

\usepackage[accepted]{icml2023}

\usepackage{amsmath}
\usepackage{amssymb}
\usepackage{mathtools}
\usepackage{amsthm}

\usepackage[capitalize,noabbrev]{cleveref}

\theoremstyle{plain}

\theoremstyle{definition}

\theoremstyle{remark}

\usepackage[textsize=tiny]{todonotes}

\begin{document}

\onecolumn
\icmltitle{HyperQuery: Beyond Binary Link Prediction}




\begin{icmlauthorlist}

\icmlauthor{Sepideh Maleki}{yyy}
\icmlauthor{Josh Vekhter}{yyy}
\icmlauthor{Keshav Pingali}{yyy}

\end{icmlauthorlist}

\icmlaffiliation{yyy}{Department of Computer Science, The University of Texas at Austin}

\icmlcorrespondingauthor{Sepideh Maleki}{smaleki@utexas.edu}

\icmlkeywords{Machine Learning, ICML}

\vskip 0.3in



\printAffiliationsAndNotice{}  

\begin{abstract}
Groups with complex set intersection relations are a natural way to model a wide array of data, from the formation of social groups to the complex protein interactions which form the basis of biological life. One approach to representing such “higher order” relationships is as a hypergraph. However, efforts to apply machine learning techniques to hypergraph structured datasets
have been limited thus far. In this paper, we address the problem of link prediction in knowledge hypergraphs as well as simple hypergraphs and develop a novel, simple, and effective optimization architecture that addresses both tasks. Additionally, we introduce a novel feature extraction technique using node level clustering and we show how integrating data from node-level labels can improve system performance. Our self-supervised approach achieves significant improvement over state of the art baselines on several hyperedge prediction and knowledge hypergraph completion benchmarks. 
\end{abstract}

\section{Introduction}

Graphs can accurately capture binary relations between entities, but they are not a natural representation of n-ary relations between entities. For example, a protein complex network cannot be represented by a graph since a protein complex might be created only in a presence of more than two proteins~\cite{crum}. 
Hypergraphs are generalization of graphs for representing such n-ary relations. 

Formally, a hypergraph $H$ is a tuple $(V,E)$ where $V$ is a set of \emph{nodes}; $E \subseteq 2^{|V|}$ is a set of nonempty subsets of $V$ called {\em hyperedges}.
 Similarly, a \textit{knowledge hypergraph} is a generalization of a knowledge graph where relations are between any number of entities. 
Recent research shows that hypergraph models produce more accurate results in problems in which graphs are used to represent n-ary relations ~\cite{hypergraph,betweeness,knowledge}. 
One important task that arises in this domain is reasoning about complex queries that go beyond a simple graph i.e. we would like to reason about a set of entities and their relations. For example, given a set of proteins, can we predict if they form a protein complex, and if so, what is the functionality of this protein complex. 

While there has been a few studies for each part of this problem, there has not been a single framework that could answer both questions. 
In this paper, we formulate the task of reasoning about hyper-queries as a \textit{higher order link prediction} problem and specifically, we propose a framework, \textit{HyperQuery}, that aims to solve the task of \textit{higher order} link prediction on both simple and knowledge hypergraphs.

Higher order link prediction in simple hypergraphs is known as hyperedge prediction and it is analogous to link prediction in graphs. Hyperedge prediction can be formally defined as follows: {\em given a hypergraph $H = (V,E)$ and a k-tuple of nodes $(v_1,v_2,...,v_k)$, predict whether this tuple forms a hyperedge or not}.
While link prediction in graphs is a well-studied problem, hyperedge prediction has not received adequate attention in spite of its many applications. For example, it can be used to predict new protein complexes, drug-drug interactions, new collaborations in citation networks, discover new chemical reactions in metabolic networks, etc.\cite{nhp,crum,disgenet}. 


\begin{figure}
     \centering
    \includegraphics[page=13,width=0.5\columnwidth]{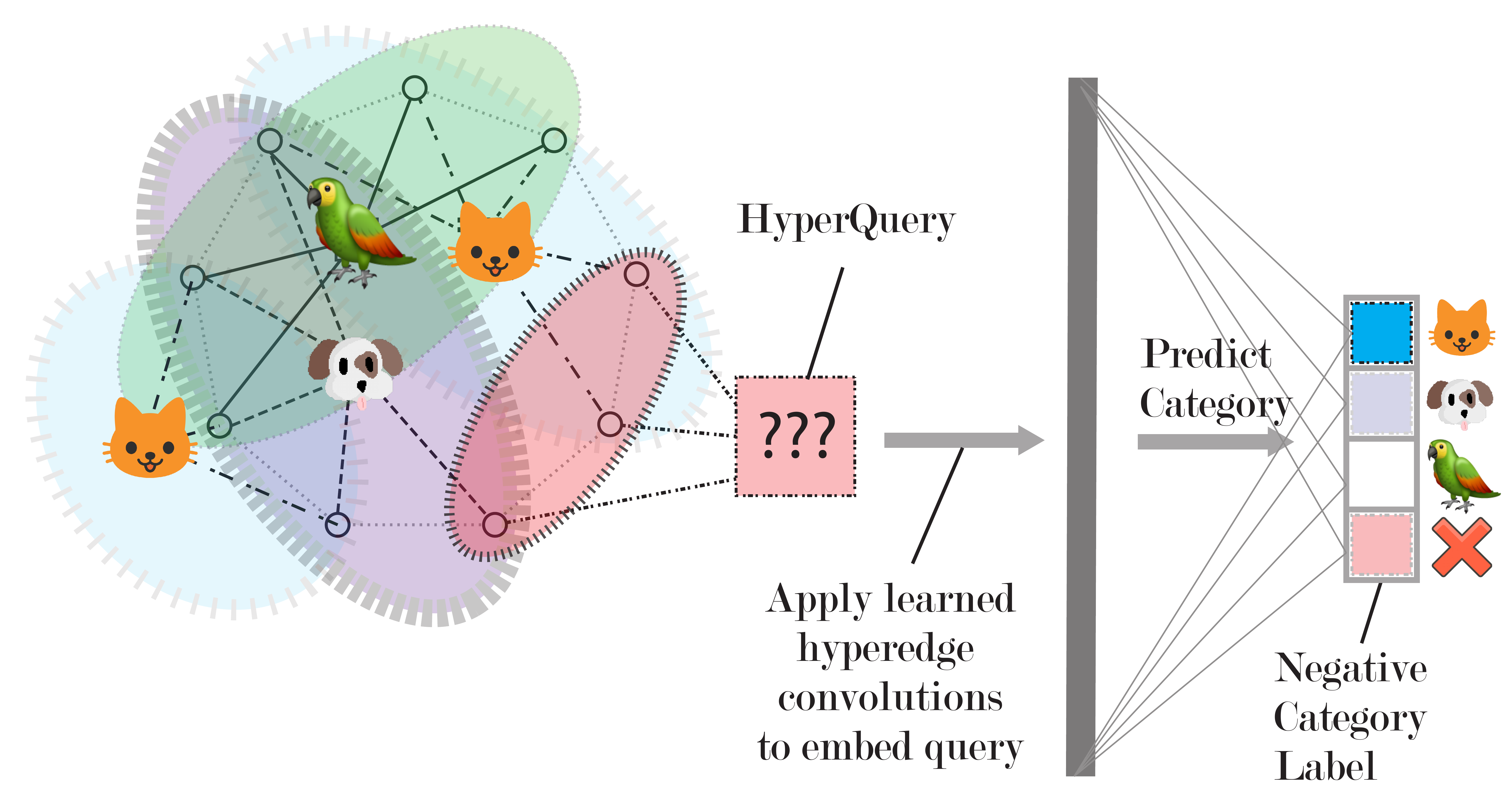}
    \caption{HyperQuery processing pipeline: Consider the hypergraph structured data illustrated above (\emph{left}).  Our system predicts the label of the red hyperedge by performing regression (\emph{right}) based on an embedding (\emph{center, grey line}) produced by the hyperedge convolution operation that we developed.
    }
    \label{fig:pipeline}
\end{figure}

In knowledge hypergraphs, it is often necessary to not only predict new hyperedges but also their relation/type/label. For example, in a protein-drug genomics knowledge hypergraph, it is important to predict not only drug-drug interactions but also the type of these interactions that describes the side effects ~\cite{decagon}. This generalized hyperedge prediction problem is also called knowledge hypergraph completion and can be described formally as follows: {\em given a knowledge hypergraph $KH = (\mathcal{E},\mathcal{R})$ and a tuple of entities $(e_1,e_2,...,e_k)$, we want to predict if this tuple forms a hyperedge and if so, what its type is}. This problem setting is illustrated in figure~\ref{fig:pipeline}.

Unfortunately, despite the fact that these problems appear to be related, the approaches to solving them differ significantly, and one cannot use the same framework to solve both problems. More precisely, hyperedge prediction requires understanding global properties of a hypergraph as well as the local features. 

Many existing learning based approaches to hyperedge prediction build systems based purely on local features like those obtained by node2vec.  We believe that bootstrapping local approaches like random walks or graph convolutions with labels which capture global structure in the graph connectivity is a broadly fruitful direction for future research.  In our specific case, we chose to obtain global features by leveraging a state-of-the-art hypergraph partitioner to cluster the graph nodes.  

In our emperical evaluations, we find that bootstrapping our (local) message-passing algorithm with global features is sufficient to achieve state of the art performance on the task of hyperedge prediction in both simple and knowledge hypergraphs. 


We summarize our contributions as follows:

\begin{enumerate}[label=(\roman*)]
    \item We develop a principled hyperedge convolution operator and develop a way to effectively apply it to both the tasks of hyperedge prediction and knowledge hypergraph completion in an \emph{entirely self-supervised} fashion.  
    \item We document an approach to obtaining an initial self-supervised embedding through hypergraph clustering has not been tried before for hypergraph completion and hyperedge prediction.
    \item We extend our neural message passing based framework to find embeddings of hyperedges in a semi-supervised fashion when some labels are already known.
    \item We evaluate our method and find that it produces state-of-the-art results on various datasets for hyperedge prediction and knowledge hypergraph completion. 
\end{enumerate}

The outline of our system is shown in figure~\ref{fig:pipeline}.

\section{Related Work}
\label{sec:related}
In this section, we review existing works for  feature generation and hyperedge prediction in simple and knowledge hypergraphs. 

\paragraph{Clustering based feature generation.} Random-walk based approaches are one way to generate features for a hypergraph, but they only exploit the local connectivity of nodes in the hypergraph. In this paper, we investigate the effectiveness of hypergraph clustering algorithms for feature extraction. 

Clustering has been used to understand the structure of complex networks. A cluster in a hypergraph is a partition of nodes into sets that are similar to each other and dissimilar from the rest of the hypergraph. Intuitively, a cluster is a group where nodes are densely inter-connected and sparsely connected to other parts of the network. Placing the embedding of such nodes closely in the embedding space can be useful in many graph mining tasks such as community detection, node classification, network reconstruction and link prediction ~\cite{louvainNE}.

A classic clustering algorithm in hypergraphs is the \textit{FM}~\cite{FM} algorithm: given an initial assignment of nodes to clusters, this algorithm moves a node to the cluster that results in the largest reduction in connectivity between clusters. Multi-level clustering approaches such as ~\cite{hmetis,phmetis,bipart} build on this algorithm and improve the performance of FM algorithm by successively coarsening the hypergraph, finding clusters in the smallest hypergraph, and then interpolating these to the coarser hypergraphs, applying the FM algorithm at each level. In this paper, we use a recent work called BiPart, \cite{bipart}, for clustering to generate initial features of the hyperedges as well as nodes of a hypergraph.

\paragraph{Random walk based feature generation for hypergraphs}
 In problems in which nodes do not have features, the first step is to generate features for the nodes. The most common approach is to use node2vec since it can easily be applied to hypergraphs. For example, HyperSAGNN ~\cite{hypersagnn} first generates features using node2vec and then passes these features through an attention layer. During inference, the embedding of a proposed tuple of nodes is computed using a one layer, position-wise feed-forward network. They can also use the corresponding row of the adjacency matrix to extract features. Similarly, NHP runs node2vec on the clique expansion of a hypergraph and then uses a GCN layer to improve these features. Finally, they pass these embeddings to a scoring layer~\cite{nhp}.
 
 There has also been a theoretically exciting line of work (surveyed in:~\cite{BATTISTON20201}) which considers various methods for extending random walks to hypergraphs.



\paragraph{Higher order link prediction in simple hypergraphs.} One way to perform link prediction in hypergraphs is by first computing the embedding of the nodes of the hypergraph and then applying a function $g$ to the tuple $(v_1,v_2,...,v_k)$ to obtain the embedding of the tuple i.e. hyperedge. A binary classifier can then be applied to these embeddings to perform hyperedge prediction.
The difficulty with such models is that first, in order to get good predictions, we need to obtain high quality embeddings. Second, for hypergraphs function $g$ must be nonlinear to capture higher-order proximity of nodes in the hypergraph~\cite{dhne}. This means it is not a good design choice to use an operator such as average to obtain the embedding of a hyperedge from its nodes. Related works HyperSAGNN and NHP have used non-linear functions such graph neural networks~\cite{hypersagnn,nhp} to compute the embedding of a hyperedge from its nodes. 

\paragraph{Higher order link prediction in knowledge hypergraphs.} This task is also known as knowledge hypergraph completion. Link prediction in knowledge hypergraphs is relatively under explored. Most approaches generalizes knowledge graph based methods to work with n-ary relations. More recent work on this area propose novel methods that directly works with knowledge hypergraphs such as ~\cite{knowledge,hkg}.


    

\begin{figure*}[h]
    \centering
    \includegraphics[page=11,width=\textwidth]{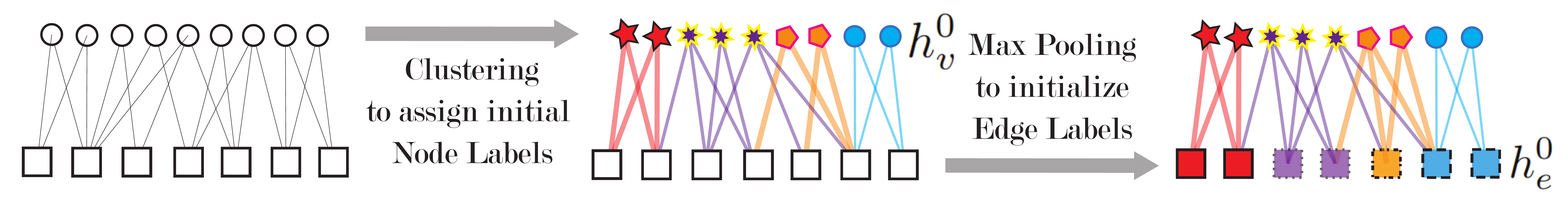}
    \caption{ Generating Useful Labels using clustering: Not all hypergraph data comes with labels (\emph{left}). %
    We leverage hypergraph clustering to first assign cluster id as labels to the nodes, obtaining an initial cluster assignment $h^0_v$ (\emph{center}), which we then propagate to the hyperedges via Max Pooling (\emph{right}) in order to obtain the initial edge cluster assignment $h^0_e$.} 
    \label{fig:labeling_from_clusters}
\end{figure*}

\section{Self-supervised Higher-order Link Prediction}
\label{sec:method}
We begin with the question of how one might build a system to ``answer'' hyper-queries. Given a collection of nodes, our HyperQuery oracle will answer questions such as whether these nodes are related. Ultimately, our goal is to perform hyperedge prediction on a collection of nodes. One main challenge towards our goal is that, in most hypergraph datasets, initial features for nodes do not exist which makes convolution operator less natural for hypergraphs.

In some cases like knowledge hypergraphs, hyperedges have types which could be used as initial features. Nodes, however, do not, resulting in the same problem.
To address the above problem, we propose a novel feature extraction method which is based on clustering nodes and hyperedges of a hypergraph. 




\subsection{Good Initial Features} 
\label{subsec:label}
In a setting where there is no additional label/feature data for nodes or hyperedges of a hypergraph, the inference must be computed entirely from the hypergraph structure.  Thus we need a way of computing a useful feature initialization. Our insight is that useful initial categorical embeddings can be found by existing clustering algorithms.  

One way to learn an embedding that captures meaningful structural information about a hypergraph is to run random-walk based algorithms such as node2vec~\cite{node2vec}. Previous works on hyperedge prediction such as HyperSAGNN~\cite{hypersagnn} and NHP~\cite{nhp} use node2vec to first learn feature vectors for nodes of a hypergraph represented in a variety of ways. Then, they improve these features using attention-based methods. However, such methods have a number of drawbacks:
\begin{enumerate}
    \item They do not learn the correlation between nodes that do not appear in the same random walk and may require many samples to capture evidence of weak correlations.
    \item Random walks explore only the local properties of the hypergraph, optimization might get stuck in local minima.  
    \item Random walks on hypergraphs mix very quickly due to high edge degree and thus even if an algorithm provably converges, computing this answer might not be practical in practice.
    \item There is no standard practice for how to transfer node level features into a feature vector per hyperedge, which we needed for the hypergraph convolution operator presented in section~\ref{sec:knowledge}.  
\end{enumerate}

\paragraph{Feature extraction using clustering}It occurred to us that instead, it is possible to use state-of-the-art partitioning based clustering tools as an alternative approach for pre-computing features.  While clustering is a well studied problem in ML, for instance it has been used in link prediction as early as in 2015~\cite{link_prediction_community_detection}, it is not used by any of the state-of-the-art prediction tools today, and therefore, we believe that this global clustering approach merits more careful study, given the simplicity of the method and effectiveness in empirical evaluations. 

In particular, for our HyperQuery flow, we want to recover an initial label assignment that is ``smooth'' with respect to the underlying hypergraph topology.  To do this, we run a clustering  algorithm that partitions the nodes of the hypergraph into clusters. Each cluster of nodes is given a unique integer id.  To transfer these labels onto hyperedges, we perform a max-pooling step (described in more detail in Figure~\ref{fig:labeling_from_clusters}).  In particular, after performing clustering, we obtain a set of features $h_e^0$ and $h_v^0$ for every node and hyperedge in the input hypergraph.  These features are then used as initialization for the graph convolution operator we describe below.


\paragraph{Cut-metric for hypergraphs.} For our purpose, we can use any clustering algorithm. In our experiments, we use a multilevel clustering algorithm that partitions nodes into a given number of clusters while minimizing the hyperedge cut, defined below. While there is not a unique canonical way to define a hyper-edge cut, this approach worked well and exposes the number of clusters as a hyper-parameter to the system. 

\begin{align}
     cut(H,C) =  \sum_{e} (\lambda_e(H,C) - 1)\label{eq:6}
\end{align}

Here, $H$ is the hypergraph, $C$ is the partition of nodes into clusters, and $\lambda_e(H,C)$ is the number of clusters that hyperedge $e$ spans. Intuitively, nodes that belong to the same cluster are considered similar. This is similar to approaches like node2vec in which nodes that appear in the same random walk are considered to be similar. 

\subsection{Hyperedge Convolution Operator}
\label{sec:knowledge}
In this section, we start by studying the question of inference in knowledge hypergraphs, where hyperedges usually have a specific type/label. We present an approach for defining trainable convolutions on hyperedges which is well suited for performing inference for knowledge hypergraph completion. 

\begin{figure}[h]
     \centering
    
    \includegraphics[width=0.6\columnwidth]{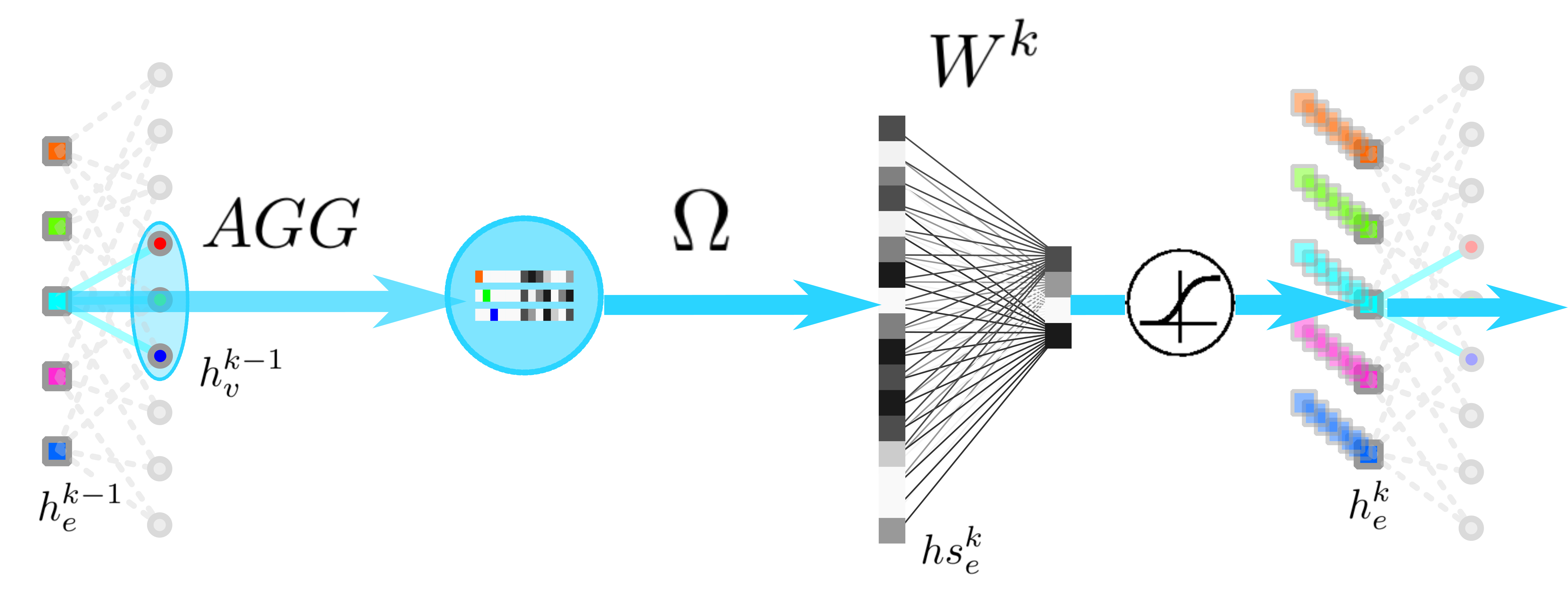}

         \caption{An Edge2Edge Convolution Operator: Here we illustrate the function $[h_v^k,h_e^k] = E2E(S, h_e^{k-1})$.
         Our query set $S$ is annotated in blue on (\emph{left}).  For each $v_i \in S$, we perform an E2N aggregation, and in the process, compute $h_v^k$  (\emph{center left}).  We then summarize the distribution of node features using some choice of function $\Omega$ to obtain a statistical feature we call $hs_e^k$ (\emph{center right}).  Passing these features through a trainable weight matrix completes the operator (\emph{right}). }
\label{fig:n2e}
\end{figure}

Our proposed hyperedge convolution operator is motivated by the intuition of trying to learn an embedding for hyperedges given their labels.  In this setting, we might imagine using some statistical measure $\Omega$ (like mean or variance) to summarize the labels in the neighborhood of each hyperedge, and then training a MLP layer based on these summary statistics, as illustrated in figure~\ref{fig:n2e}. 

Our convolution operator is based on a message passing framework where messages are passed between nodes and hyperedges alternatively and final hyperedge representations are pooled from node embeddings. 
In particular, our framework consist of two operators: we have an \emph{edge2node} convolution operator which is formulated as a harmonic mean of neighboring edge embeddings using the function $AGG$ followed by a \emph{node2edge} convolution operator.

\begin{align}
E2N(v_i,h_e^{k-1}) &= h_{v_i}^k= AGG({h_e^{k-1}, \forall e \in \mathcal{N}(v_i)}) \label{eq:e2n} \\
N2E(S,h_v^{k}) &= h_S^k = \sigma(W^k \cdot \Omega \{h_v^{k-1}\forall v \in S\}) \label{eq:n2e}
\end{align}

Equation \ref{eq:e2n} defines an \emph{edge2node} operator where $v_i$ is a node whose neighboring hyperedge features we want to aggregate. While equation \ref{eq:n2e} defines a \emph{node2edge} convolution operator, as long as we take $S = \mathcal{N}(e)$ (neighborhood of hyperedge $e$) for a known hyperedge $e$.  That said, we can perform $N2E$ summarization on sets of nodes which are not known to share a hyperedge, and in fact it can just as well be thought of as .  In fact, this idea will enable our formulation to generalize to unknown data after appropriately training the weight matrices $W^k$ and nonlinear activation $\sigma$.

By combining these two operators, we arrive at the Edge2Edge convolution operator which is defined below, and illustrated more carefully in figure~\ref{fig:n2e}.
\begin{align}
E2E(S,h_e^{k}) = N2E(S, \{\forall v_i \in S, E2N(v_i,h_e^{k}) \} ) \label{eq:e2e}
\end{align}

This approach to obtaining hyperedge embeddings is natural and simple, but the challenge is that it depends on there existing high quality node level data $h_v$.  Our clustering based feature extraction framework provides such high quality node/hyperedge data and thus the two techniques complement each other.

\paragraph{Node cluster as a feature.}
The initial aggregation step $AGG$ in our convolution operator comes from the idea that a reasonable first way to think about creating embeddings for hyperedges is to simply average node embeddings.  However, this simple measure fails to capture important information regarding the distribution of node embeddings within a hyperedge.  
To fix this problem, we augment every node feature vector with the a one-hot encoding of the node cluster id, as seen in Equation~\ref{eq:concat}:
\begin{align}
h_{v}^k &= CONCAT\{ AGG\{{h_e^{k-1}, \forall e \in \mathcal{N}(v)}\}, x_v\} \label{eq:concat}
\end{align}
where $AGG(.)$ is an aggregation function, and $x_v$ is the one-hot partition id of a node. If nodes of a hypergraph have features themselves, these features can also be concatenated with $x_v$.


 
\subsection{Optimizing HyperQuery}

Here we outline a few important technical details that significantly improve our framework.

\paragraph{Choosing the right $\Omega$.}
\label{subsec:omega}
In this work, we evaluate three different types of summary statistics for $\Omega$.  A component-wise mean, a component-wise variance, and a sort of ``robust variance'' estimator we call \textit{minmax} (element-wise difference of maximum and the minimum values of the vectors) similar to the one used in NHP~\cite{nhp} to achieve state of the art performance.  Intuitively, when $\Omega$ is the mean, convolution maps hyperedges to the average embedding of constituent nodes, whereas the variance measures how correlated different nodes are with each other.  In our experiments in Section \ref{sec:exp}, we find that for the hyperedge prediction, \textit{minmax/variance} performed better as a choice of $\Omega$ whereas in knowledge hypergraph completion, \emph{mean} worked better.

\paragraph{Bilinear aggregation.}
The final optimization that we use in our scheme to better capture the correlation between different labels/communities is inspired by a bilinear aggregation layer~\cite{bilinear}.. Bilinear aggregation has significantly improved visual concepts recognition~\cite{bilinear2}, and in practice also generally improves the accuracy of our results. Specifically, \textit{we create an auto-correlation matrix by multiplying a hyperedge feature vector by itself}. This matrix has dimension $d \times d$, which we then flatten.

For our final scheme for hyperedge message passing, we rewrite Equation \ref{eq:e2e} as the following:
\begin{align}
\Omega^k_e &= \Omega\{h_v^k ,\forall v \in \mathcal{N}(e)\} \label{eq:8}\\ 
h_e^k &= \sigma(W^k \cdot  flat(\Omega^k_e \Omega^{k^T}_e)) \label{eq:9} 
\end{align}

Where $h_{v}^k$ is computed using Equation~\ref{eq:concat}, $\Omega$ is an aggregation function, $flat(.)$ flattens the correlation matrix and stores it in a column vector, 
and $W^k$ is the trainable weight matrix. The effect of using different aggregation functions $\Omega$ is explored empirically in Table~\ref{tb:prediction}.



\subsection{Knowledge Hypergraph Completion}
In a knowledge hypergraph, we can effectively learn the hidden weights $W^k$ by choosing the initial feature $h^0_e$ of a hyperedge $h_e$ to be the one hot vector of its type concatenated by its cluster id, and building up an auto-encoder style problem.  
By repeating this iteration twice, we already arrive at an effective tool for knowledge hypergraph completion which generalizes to unseen data far better than benchmark approaches.  In the experiments, we outline our performance relative to other benchmarks.  We achieve this performance with a straightforward model where we simply treat the output $h_e^2$ as a predicted class label $\mathcal{P}(e)$.  From there we minimize the loss below as a function of hidden layer weights $W^1$ and $W^2$.  
\begin{align}
 \mathcal{L} = \sum_{e \in \mathcal{D}} \mathcal{J}( \mathcal{P}(e), \mathcal{G}(e)) \label{eq:5}
\end{align}

In our objective function, we take $\mathcal{J}(.)$ to be the cross entropy loss, $e$ to be a hyperedge in the training dataset $\mathcal{D}$, $\mathcal{P}(e)$ is the predicted type of hyperedge $e$, and $\mathcal{G}(e)$ is the actual ground truth type (relation) of hyperedge $e$ in the dataset. 


\subsection{Hyperedge Prediction}

The second problem studied in this paper is \textit{hyperedge prediction}: given a hypergraph and set of nodes $(v_1,..,v_n)$,  predict whether this tuple forms a hyperedge. The key difference between this problem and knowledge hypergraph competion is that in the latter problem, the hyperedge is known to exist and has a type/label. 

We can perform hyperedge prediction using the same framework, HyperQuery.  Once trained, this scheme can be used as a pretext task to generate the representation of the hyperedges. We then apply a fully connected layer on them to generate logits to train cross entropy loss. We train the linear classifier using the frozen hyperedge representations. 

Training is performed with actual hyperedges as well as with negative hyperedges ({\em i.e.}, node tuples that do not constitute hyperedges) that are assigned a label that is distinct from the labels in the dataset. We use a similar approach to NHP~\cite{nhp} to generate negative samples, i.e. for each hyperedge $e$, we create a negative hyperedge $\Tilde{e}$ by having half of nodes sampled from $e$ and the remaining from $V - e$. This sampling method is inspired by chemical reaction datasets. In Section \ref{sec:exp} we further explain our negative sampling strategy.

\subsection{Time Complexity} 
We analyze the time complexity of our framework (Equations \ref{eq:concat}, \ref{eq:8},and \ref{eq:9}) in terms of the size of the hypergraph, assuming the number of relations (types) is independent of the size of the hypergraph. Let $m$ denote the number of hyperedges and $n$ denote the number of nodes. Let $deg (v_i)$ denote the degree of node $v_i$ and let $\Tilde{\Delta}$ denote $max_{1\leq i\leq n} deg (v_i)$. For Equation \ref{eq:concat}, our framework takes  $O({n\cdot \Tilde{\Delta}})$ time.
For Equation~\ref{eq:8}, let $deg (e_i)$ denote the degree of hyperedge $e_i$ and let $\Delta$ denote $max_{1\leq i\leq m} deg (e_i)$.  Equation~\ref{eq:8} takes $O({m\cdot \Delta})$ time. Finally, the time complexity of Equation~\ref{eq:9} is constant since the number of labels is independent of the size of the graph. Overall, our framework takes $O({m\cdot \Delta}) + O({n\cdot \Tilde{\Delta}})$.
\subsection{Model Explainability}


The ideal initial feature vector for a hyperedge should include local and global properties of the hypergraph. In our model, clustering exposes global properties. Specifically, we keep track of the cluster id of hyperedges as well as the distribution of labels on the nodes in the neighborhood of a hyperedge. Intuitively, Hyperedges that are in the same cluster or have similar label distribution on their nodes are similar to each other and by assumption, they are related. Previous work ~\cite{node2vec, deepwalk} has shown that placing such similar nodes closely in the embedding space will facilitate tasks such as node classification and link prediction. We follow the same intuition. 

The message-passing scheme of labels is conceptually similar to label propagation~\cite{lpa}. The objective of the label propagation algorithm is to assign each node into a cluster with the most number of its neighbouring nodes. Intuitively, this scheme also put hyperedges with the same label/cluster closer in the embedding space facilitating hyperedge prediction for both knowledge hypergraphs and simple hypergraphs.

\section{Experiments}
\label{sec:exp}
We evaluate our framework HyperQuery on knowledge hypergraph completion task as well as hyperedge prediction and show that our architecture outperforms the state of the arts for these tasks. Furthermore, we conduct an ablation study to understand the effectiveness of the major HyperQuery kernels.

\paragraph{Hyper-parameters.} The number of clusters used in the evaluation section of this paper is 16. Furthermore, we showed the effect of using different number of clusters in Figure~\ref{fig:partition_depedence}. We use 2 layers of hyperedge convolution in our test implementation, and ReLU as nonlinear activation function. For the baselines, we adopt the same setting as the one used in the original papers.

\paragraph{Clustering algorithm.}
 In this paper we use BiPart~\cite{bipart} as our clustering algorithm. BiPart is a multilevel and deterministic hypergraph partitioner. Given a number $k$, BiPart partitions the nodes of the hypergraph into k disjoint blocks. In our framework, we use BiPart as a pre-processing step where we partition the nodes of a hypergraph first and the we use the one hot vector of their partition id as the initial feature of the nodes of the hypergraph. Finally, these initial features are used as an input to the HyperQuery framework. 
 
 In practice, any hypergraph partitioning algorithm can be used to partition the nodes of the hypergraph.

\begin{table*}[htbp]
\caption{Knowledge Hyperedge Completion.}
\label{tb:class}
\vskip 0.15in
\begin{center}
\begin{small}
\begin{sc}
\begin{tabular}{lccccccccccccc}
\toprule
\textbf{} &  \multicolumn{3}{c}{\textbf{FB-AUTO}}   &
 \multicolumn{3}{c} {\textbf{M-FB15K}} & \multicolumn{3}{c}{\textbf{JF17K}} \\ 
  &
  \multicolumn{1}{l}{MRR} &
  \multicolumn{1}{l}{\textit{Hit@1}} &
  \multicolumn{1}{l}{\textit{Hit@3}} &
 \multicolumn{1}{l}{MRR} &
 \multicolumn{1}{l}{\textit{Hit@1}} &
  \multicolumn{1}{l}{\textit{Hit@3}}&
  \multicolumn{1}{l}{MRR} &
  \multicolumn{1}{l}{\textit{Hit@1}} &
  \multicolumn{1}{l}{\textit{Hit@3}}\\

\midrule

HyperQuery & \textbf{91.5} & \textbf{83.1} & \textbf{99.9} & \textbf{95.1} & \textbf{90.5} & \textbf{99.7}&\textbf{99.9}&\textbf{99.1}&\textbf{99.9}\\

HSimplE& 79.8 &  76.6 & 82.1&   73.0& 66.4  &76.3 &47.2&37.8& 52.0\\
HypE& 80.4 &77.4 & 82.3  & 77.7 & 72.5 &80.0&49.4&40.8&53.8  \\
m-DistMult& 78.4 &74.5  & 81.5  & 70.5 & 63.3 &74.0&46.3&37.2&51.0 \\
m-TransH & 72.8 &72.7  & 72.8  & 62.3 & 53.1 &66.9&44.4&37.0&47.5\\
m-CP & 75.2 &70.4  & 78.5  & 68.0 & 60.5 &71.5& 39.1&29.8&44.3\\

\bottomrule
\end{tabular}
\end{sc}
\end{small}
\end{center}
\vskip -0.1in
\end{table*}

\subsection{Knowledge Hypergraph Completion}

We first evaluate our system for the task of link prediction for knowledge hyperedges i.e. knowledge hypergraph completion. We evaluate our system on three datasets: FB-AUTO, M-FB15K, and JF17K. We used the same test, train, and validation sets as \cite{knowledge}. We use MRR (mean reciprocal rank) and Hit@1, 3 (hit ratio with cut-off values of 1 and 3) as our evaluation metrics.

\paragraph{Datasets:} The datasets used in this section are standard knowledge hypergraph datasets from previous work~\cite{knowledge}. The summary of the knowledge hypergraphs are in Table~\ref{dataset_k}. FB-AUTO, M-FB15K~\cite{knowledge}, and JF17K~\cite{jf17k} are knowledge hypergraphs with n-ary relations that is collected from Freebase dataset.

\begin{table}[htbp]
\caption{Knowledge hypergraph dataset}
\label{dataset_k}
\begin{center}
\begin{small}
\begin{sc}
\begin{tabular}{lcccccc}
\toprule
Data set & $|\mathcal{E}|$ & $|\mathcal{R}|$ & \textit{\#train}& \textit{\#valid}& \textit{\#test}\\
\midrule
FB-AUTO  & 3,410& 8& 6,778 & 2,255 & 2,180\\
M-FB15K & 10,314& 71 & 415,375&39,348&38,797 \\
JF17K & 29,177 & 327 & 77,733& 15,822 &24,915  \\

\bottomrule
\end{tabular}
\end{sc}
\end{small}
\end{center}
\end{table}

We compare our results with the following baselines: \textbf{HSimplE, and HypE} two embedding based approaches for knowledge hypergraphs introduced by~\cite{knowledge}. Both methods find the embedding of an entity based on its position in a relation.  \textbf{M-DistMult}~\cite{distmul} defines a scoring function for each tuple $(e_1,r_1,e_2)$. We modify this so that each relation could have any number of entities. \textbf{M-TransH} is a modified standard knowledge graph scoring function (TranshH) that accepts beyond binary relations. 
Finally, \textbf{M-CP}~\cite{mcp} is a tensor decomposition approach. We apply a similar approach as~\cite{knowledge} to extend it beyond binary relations.

We used HyperQuery to perform knowledge hypergraph completion. Table~\ref{tb:class} summarizes the result of our experiment. HyperQuery performs the best compared to all other baselines on all metrics. Specifically, on metric Hit@1 against the best baseline, it improves up to $5\%$ on dataset FB-AUTO, $18\%$ on M-FB15K, and more than $50\%$ on JF17K.

\subsection{Hyperedge Prediction}

The second task we study in this paper is hyperedge prediction. We evaluate our system on four datasets: iAF1260b, iJO1366, USPTO, and DBLP. We used $70\%$ of the hyperedges in these for test, $10\%$ for validation and $20\%$ for training.
 The summary of these datasets are in Table~\ref{dataset}.
\begin{table}[htbp]
\caption{Hyperedge prediction dataset}
\label{dataset}
\begin{center}
\begin{small}
\begin{sc}
\begin{tabular}{lcccc}
\toprule
Data set & Nodes & Hyperedges & Type of data\\
\midrule
iAF1260\textit{b}    & 1,668& 2,084& \textit{Metabolic reactions}\\
iJO1366 & 1,805& 2,253 & \textit{Metabolic reactions}\\
USPTO & 16,293 & 11,433 & \textit{Organic reactions}  \\
DBLP    &20,685& 30,956 & \textit{Co-authorship}\\

\bottomrule
\end{tabular}
\end{sc}
\end{small}
\end{center}
\end{table}

\textbf{iAF1260b}\footnotemark[1] a metabolic reaction dataset for specie E. coli. We use this dataset for hyperedge prediction where our goal is to predict missing reactions i.e. hypereges. In this dataset each reaction is considered as a hyperedge connecting its participating metabolites (nodes).

\textbf{iJO1366} \footnotemark[1] a metabolic reaction dataset similar to iAF1260b.

\textbf{USPTO} \footnotemark[2] a organic reaction dataset. We used a subset of chemical substances that only contains carbon, hydrogen, nitrogen, oxygen, phosporous, and sulphur.

\textbf{DBLP} \footnotemark[3] a co-authorship publication dataset. We used a subset of papers published in only AI conferences: AAAI, IJCAI, NeurIPS, ICML, CVPR, ICCV, ACL, NAACL,
etc. Each author in this dataset is a node and papers represent hyperedges connecting authors of a paper.

\footnotetext[1]{ \url{https://github.com/muhanzhang/HyperLinkPrediction} }
\footnotetext[2]{ \url{https://github.com/wengong-jin/nips17-rexgen} }
\footnotetext[3]{ \url{https://github.com/muhanzhang/HyperLinkPrediction} }

We compare our results with previous works: NHP~\cite{nhp}, HyperSAGNN~\cite{hypersagnn}, HyperGCN~\cite{hypergcn}, and node2vec~\cite{node2vec}. The description of these methods are in \ref{sec:related}.

\begin{table*}[htbp]
\caption{Area Under Curve (AUC) scores for hyperedge prediction. NP refers to the model without bilinear pooling. MN refers to the minmax aggregation model. HQ is our proposed model.}
\label{tb:prediction}
\vskip 0.15in
\begin{center}
\begin{small}
\begin{sc}
\begin{tabular}{lcccccccccc}
\toprule
\textbf{} & {\textbf{iAF1260b}}   &
 {\textbf{iJO1366}}&{\textbf{USPTO}} &{\textbf{DBLP}} \\

\midrule

HQ-mn & 72.2 & 68.5  & \textbf{75.7}   & \textbf{72.0}  \\
HQ-mean & 66.5 & 65.9 & 72.2   & 69.6  \\
HQ-var & 71.1 & 68.1 & 75.0   & 71.3  \\
\midrule
HQ-mn-np & \textbf{72.7} & \textbf{68.9}  & 70.1   & 70.4  \\
HQ-mean-np & 67.6 & 65.7 & 70.6   & 65.7  \\
HQ-var-np & 72.3 & 67.5 & 74.1   & 71.6  \\
\midrule
NHP-mn & 64.3   & 63.2    &  74.2  & 69.2   \\
NHP-mean & 60.5   & 61.2    &  65.5  & 56.4   \\
Hyper-SAGNN& 60.1   & 56.3 &   67.1    & 65.2   \\
HyperGCN-mn & 64.0   & 62.2    &  70.5  & 67.4   \\
node2vec-mn& 66.0   & 62.0  &   71.0   & 67.0   \\

\bottomrule
\end{tabular}
\end{sc}
\end{small}
\end{center}
\end{table*}

\paragraph{Operator $\Omega.$} Motivated by NHP, we investigate an aggregation function $\Omega$ for our approach. We call this aggregation function \textit{minmax} which is the element wise difference of max and min values of the embedding vectors in equation. We fix the embedding dimension to 64. Furthermore, we also experiment on using operator \textit{variance} as an aggregation function. Furthermore, for a given set of vectors $x_1, ..., x_k \in \mathcal{R}^d$, 
\begin{align*}
    MINMAX(x_1,...,x_k) = (max \: x_{sl} - min \: x_{sl}), s \in [k], \notag \\
    l=1,...,d\\
    VAR(x_1,...,x_k) = (variance(x_{sl})), s \in [k], l=1,..., d
    \\
    MEAN(x_1,...,x_k) = (MEAN(x_{sl})), s \in [k], l=1,..., d 
\end{align*}

We use 16 communities for this experiment. In Section \ref{sec:abi}, we discuss the effect of different number of communities. 

\paragraph{Negative sampling.} We use a similar approach to ~\cite{nhp} i.e for each hyperedge $e$ in our dataset, we create a hyperedge $e^{\prime}$ by having half of the vertices sampled from $e$ the remaining half from $V - e$. This sampling method is motivated by the chemical reaction datasets where it is unlikely that half of the substances of a valid reaction (from $e$) and
randomly sampled substances (from $V - e$) are involved in another valid reaction. 

HyperQuery with minmax aggregation function performs the best on all datasets in this paper. Comparing with NHP, on their minmax operator, HyperQuery minmax outperform them by up to $7\%$ on iAF1260b, $5\%$ on iJO1366, $1.5\%$ on USPTO, and $1.8\%$ on DBLP; on mean aggregation operator, HyperQuery outperforms NHP by up to $6\%$ on iAF1260b, $4.6\%$ on iJO1366, $12\%$ on USPTO, and $13.6\%$ on DBLP. 
HyperQuery outperforms HyperSAGNN on all datasets by more than $6\%$, and HyperGCN by more than $4\%$. Finally, node2vec results show poor performance on hypergraph datasets showing that graph based approaches are not suitable for hypergraphs.

\subsection{Ablation Study}
\label{sec:abi}
We study the behaviour of each component of HyperQuery. 
First, we study the effect of bilinear pooling on hyperedge prediction. Table~\ref{tb:prediction}, shows this effect. On the larger dataset, USPTO, the bilinear pooling improves the quality of our framework by $5.6\%$ for minmax. Similarly, for DBLP, the improvement is about $2\%$. On the smaller datasets, bilinear pooling does not make a significant change.

\begin{figure}
     \centering
    \includegraphics[width=0.6\columnwidth]{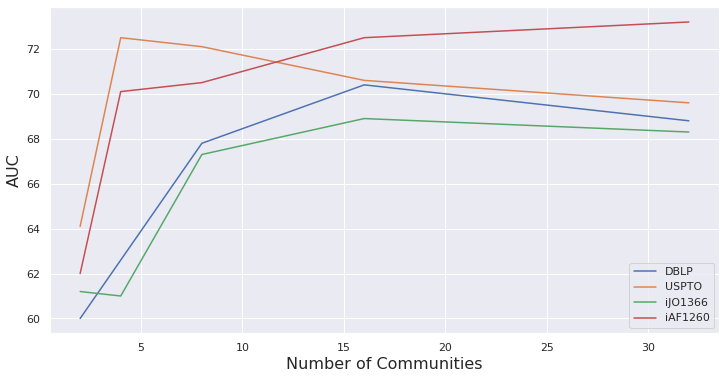}
    \caption{Performance of HyperQuery for different numbers of clusters.}
    \label{fig:partition_depedence} 
    
\end{figure}

Next, we study the effect of the number of clusters in our dataset. Figure~\ref{fig:partition_depedence} shows how the performance of our framework changes as we increase or decrease the number of clusters. For all datasets except iJO1366, the performance improves as we increase the number of clusters up to 16 clusters and then decreases. This behaviour is expected since as we increase the number of clusters, the quality of these clusters decreases ({\em i.e.}, the hyperedge cut increases). However, if the number of clusters is very small, we are not exploring the global structure of our hypergraph enough which again decreases the performance of our framework.

\section{Conclusion and Future Work}
In this paper, we introduced a neural message passing based framework, HyperQuery, that could handle two tasks: knowledge hypergraph completion and hyperedge prediction. We designed two convolution operators that pass messages between nodes and hyperedges alternatively and final hyperedge representations are pooled from node embeddings.
Our framework utilizes clustering results on a hypergraph to further improve and solve the prediction problem on hyperedges. This novel feature extraction method improves our framework by augmenting global information of the hypergraph. In our experiments, we demonstrate that HyperQuery outperforms state-of-the-art approaches in both knowledge hypergraph completion and hyperedge prediction in simple hypergraphs. 

In future, we would like to answer more complicated queries using HyperQuery and combine different statistics of the hypergraph to improve the generated embeddings.

\nocite{langley00}

\bibliography{main}
\bibliographystyle{icml2023}

\newpage
\appendix
\onecolumn





\section{Choosing the Right $\Omega$}
\begin{figure}[h]
\centering
    \includegraphics[page=16,width=\textwidth]{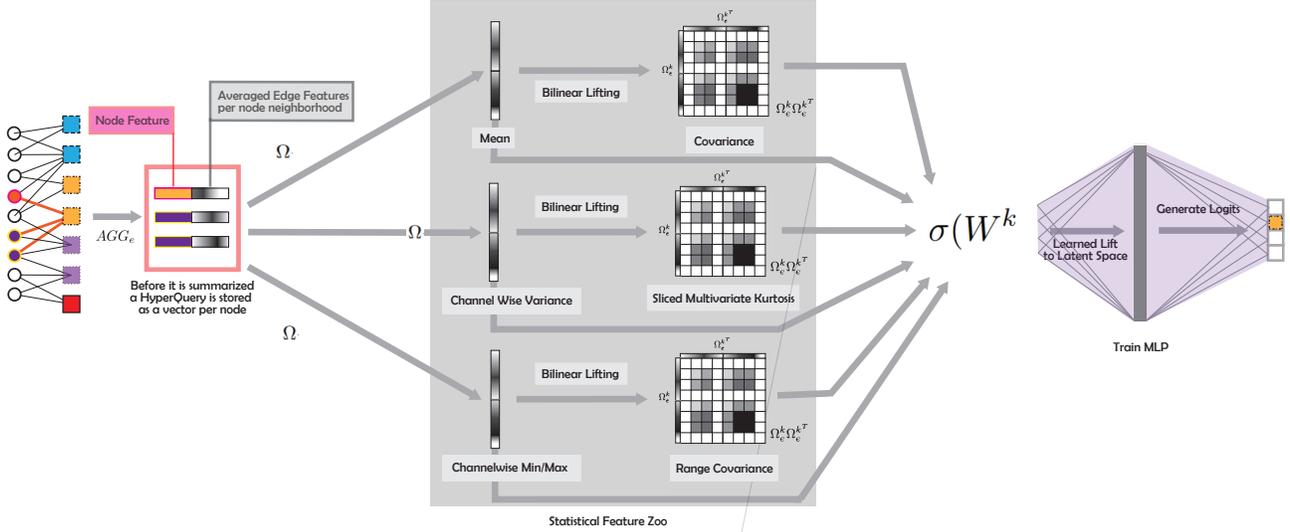}
    \caption{ Learning $W^1$.  Given a hypergraph with labels on nodes and edges, we experiment with a zoo of choices for $\Omega$.  In this illustration, we convolve once, multiply by $W^1$ and $\sigma$ and then pass through a fully connected layer to return to a space of the same dimension of hyperedge types.  Finally we then train as an autoencoder.}

    \label{fig:zoo}
\end{figure}

In this work, we evaluate three different types of summary statistics for $\Omega$ (Figure~\ref{fig:zoo}, center).  A component-wise mean and variance, and a sort of ``robust variance'' estimator we call \textit{minmax} (element-wise difference of maximum and the minimum values of the vectors) similar to the one used in NHP~\cite{nhp} to achieve state of the art performance.  Intuitively, when $\Omega$ is the mean, convolution maps hyperedges to the average embedding of constituent nodes, whereas the variance measures how correlated different nodes are with each other.  In our experiments in Section 6, we find that for the hyperedge prediction, \textit{minmax/variance} performed better as a choice of $\Omega$ whereas in knowledge hypergraph completion, \emph{mean} worked better.  

\section{Hyperedge Classification}
One neutral choice for our system would be to use it to solve hyperedge classification. In such problems, a hypergraph $H$ is a tuple $(V,E,\mathcal{L})$ where $V$ is a set of \emph{nodes}; $E \subseteq 2^{|V|}$ is a set of nonempty subsets of $V$ called {\em hyperedges}; and $\mathcal{L}$ is a set of labels for hyperedges. Our task is given a hypergraph and labels on a small subset of hyperedges, predict labels on the remaining hyperedges.

We evaluate our system on three datasets: Cora, Citeseer, and Pubmed~\ref{aap_dataset}. We used $20\%$ of the hyperedges in these datasets for test, $10\%$ for validation and $70\%$ for training (for 20 epochs).
Since there are no previous approaches for hyperedge classification, we modified existing methods as baselines for our problem:

\textbf{HyperNetVec~\cite{netvec}:} an unsupervised multi-level approach to generate the representation of a hypergraph. This method uses an existing embedding system to generate an initial embedding and further improve these embeddings using a refinement algorithm. This methods generates embeddings for nodes of a hypergraph. We modified this method to also generate embeddings for hyperedges.

\textbf{node2vec(mean):} a random walk based approach to generate representation of a graph in a semi-supervised manner. In order to use node2vec, we convert our hypergraph to a graph and then generate embeddings for the nodes of the hypergraph. We obtain embedding of a hyperedge by performing a mean aggregation on the embedding of its nodes. 

We used HyperQuery to run these experiments. Table~\ref{tb:class} summarizes the result of our experiment. HyperQuery performs the best compared to HyperNetVec and node2vec. It improves up to $2.3\%$ on dataset Cora, $18\%$ on Citeseer, and $4\%$ on Pubmed.

\begin{table*}[htbp]
\caption{Real world hypergraph dataset for classification}
\label{aap_dataset}
\begin{center}
\begin{small}
\begin{sc}
\begin{tabular}{lccccr}
\toprule
Data set & Nodes & Hyperedges & Type of data & Classes\\
\midrule
Cora    & 2,709& 1,963& citation & 7 \\
Citeseer & 3,328& 2,182& co-authorship &6\\
PubMed    & 19,717& 12,971& co-authorship& 3 \\

\bottomrule
\end{tabular}
\end{sc}
\end{small}
\end{center}
\end{table*}

\begin{table*}[htbp]
\caption{Hyperedge classification. Accuracy in $\%$ and time in seconds.}
\label{tb:class}
\begin{center}
\begin{small}
\begin{sc}
\begin{tabular}{lcccccccccc}
\toprule
\textbf{} &  \multicolumn{2}{c}{\textbf{Cora}}   &
 \multicolumn{2}{c} {\textbf{Citeseer}} & \multicolumn{2}{c}{\textbf{Pubmed}} \\ 
  &
  \multicolumn{1}{l}{Accuracy} &
  \multicolumn{1}{l}{Time} &
 \multicolumn{1}{l}{Accuracy} &
 \multicolumn{1}{l}{Time} &
  \multicolumn{1}{l}{Accuracy}&
  \multicolumn{1}{l}{Time} &\\

\midrule

HyperQuery-mean & \textbf{78.5} & 10 & \textbf{75.3} & 12 & \textbf{82.4} & 25\\
HyperNetVec& 76.2&  28 & 57.1&   45     & 78.1  &33 \\
node2vec-mean& 76.0 &30  & 57.1  & 31     & 74.2 &140  \\
\bottomrule
\end{tabular}
\end{sc}
\end{small}
\end{center}
\vskip -0.1in
\end{table*}


\end{document}